\documentclass[journal]{IEEEtran}
\ifCLASSOPTIONcompsoc
\usepackage[nocompress]{cite}
\else
\usepackage{cite}
\fi
\ifCLASSINFOpdf
\else
\fi
\usepackage{algpseudocode}
\usepackage{amsmath}
\usepackage{algorithmicx}
\usepackage{algorithm}
\usepackage{graphicx}
\usepackage{amssymb,subfigure,cases}
\usepackage{times}
\usepackage{color}
\usepackage{multirow}
\usepackage{array}

\begin{document}
	
\title{EA-Net: Edge-Aware Network for Flow-based Video Frame Interpolation}
	
	\author{Bin~Zhao and
		Xuelong~Li,~\IEEEmembership{Fellow,~IEEE}
		\IEEEcompsocitemizethanks{
			\IEEEcompsocthanksitem This work is supported by The National Key Research and Development Program of China (No. 2018AAA0102201), by The National Natural Science Foundation of China (Nos. 61871470, 61761130079, U1801262), by Key Research Program of Frontier Sciences, Chinese Academy of Sciences (No. QYZDY-SSW-JSC044). \emph{(Corresponding author: Xuelong Li.)}			
			
			\IEEEcompsocthanksitem Bin Zhao is with School of Computer Science and Center for OPTical IMagery Analysis and Learning (OPTIMAL), Northwestern Polytechnical University, Xi'an 710072, Shaanxi, P. R. China (binzhao111@gmail.com).
			
			\IEEEcompsocthanksitem Xuelong Li is with School of Computer Science and Center for OPTical IMagery Analysis and Learning (OPTIMAL), Northwestern Polytechnical University, Xi'an 710072, Shaanxi, P. R. China (xuelong\_li@nwpu.edu.cn).

		
	}
		
	}
\IEEEtitleabstractindextext{%
\begin{abstract}
Video frame interpolation can up-convert the frame rate and enhance the video quality. In recent years, although the interpolation performance has achieved great success, image blur usually occurs at the object boundaries owing to the large motion. It has been a long-standing problem, and has not been addressed yet. In this paper, we propose to reduce the image blur and get the clear shape of objects by preserving the edges in the interpolated frames. To this end, the proposed Edge-Aware Network (EA-Net) integrates the edge information into the frame interpolation task. It follows an end-to-end architecture and can be separated into two stages, \emph{i.e.},  edge-guided flow estimation and edge-protected frame synthesis. Specifically, in the flow estimation stage, three edge-aware mechanisms are developed to emphasize the frame edges in estimating flow maps, so that the edge-maps are taken as the auxiliary information to provide more guidance to boost the flow accuracy. In the frame synthesis stage, the flow refinement module is designed to refine the flow map, and the attention module is carried out to adaptively focus on the bidirectional flow maps when synthesizing the intermediate frames. Furthermore, the frame and edge discriminators are adopted to conduct the adversarial training strategy, so as to enhance the reality and clarity of synthesized frames. Experiments on three benchmarks, including Vimeo90k, UCF101 for single-frame interpolation and Adobe240-fps for multi-frame interpolation, have demonstrated the superiority of the proposed EA-Net for the video frame interpolation task.
\end{abstract}

}

\maketitle
\IEEEdisplaynontitleabstractindextext

\IEEEpeerreviewmaketitle

\ifCLASSOPTIONcompsoc
\IEEEraisesectionheading{\section{Introduction}\label{sec:introduction}}
\else

\section{Introduction}

With the popularity of smartphones and digital cameras, people are full of desire to record and share their memorable moments in daily lives \cite{li2017locality,chen2019temporally}. However, the quality of user captured videos are usually uneven, which impacts the visual effect. In this case, there is an urgent demand for automatic techniques to enhance the video quality \cite{tian2020tdan,li2017multiview}. Video frame interpolation is such an effective technique rising for video quality enhancement.

Video frame interpolation can up-convert the frame rate of videos by interpolating intermediate frames between consecutive frames \cite{shen2020blurry,DBLP:journals/tmm/UsmanHLXBC16}, \emph{e.g.}, from 30-fps (frames per second) to 240-fps, so that the motions in the video stream become smoother. By taking this advantage, it can be applied to different video processing tasks, such as animation production \cite{pla2019computer,wang2019learning}, high-speed photography \cite{deng2019sinusoidal,jiali2020multichannel}, slow motion generation \cite{xiang2020zooming,funatsu20198k}, and so on. Overall, video frame interpolation is a fundamental task in computer vision and graphics with great application potentials. 
 
 Generally, there is a stable development of video frame interpolation in the literature \cite{DBLP:journals/spic/TuXZPVLY19,zhang2019survey}. Traditional approaches are mainly composed of two parts, \emph{i.e.}, motion estimation and frame synthesis \cite{ha2004motion,huang2008multistage}. They estimate the object motion by computing the optical flow between consecutive frames, and synthesize the intermediate frames by warping the two frames according to the computed flow map. Recently, video frame interpolation has achieved great progress benefiting from the Convolutional Neural Networks (CNNs) \cite{lee2020adacof,niklaus2020softmax,choi2020channel}. The CNN-based approaches share similar architectures with traditional ones. The main differences lie in that CNNs are adopted for flow estimation and frame synthesis by taking advantages of their visual feature capturing ability. Furthermore, the flow estimation and frame synthesis parts are combined together in CNN-based approaches for end-to-end training.
 
\subsection{Motivation and Overview}
 
Although tremendous progress has been made in video frame interpolation, it is still a challenging task for high-quality frame generation. In general, large motion and occlusions of objects tend to cause frame blur and artifacts. Existing approaches try to address this problem by employing more advanced flow estimators pre-trained on other large-scale datasets \cite{DBLP:conf/iccv/DosovitskiyFIHH15,sun2018pwc}. Moreover, the depth information and object masks are also employed to predict the occlusions \cite{bao2019depth,bao2019memc}. By integrating more information, the performance is improved. However, these approaches are in more complex architectures and require more pre-trained models and extra annotated datasets, which increases the training difficulty and makes them hard to operate in the wild.

Practically, the frame blur and artifacts caused by large motion and occlusion usually occur at the object boundaries. Considering that frame edges draw the boundaries of different objects, we propose to improve the quality of interpolated frames by preserving the frame edges explicitly. Motivated by this, the Edge-Aware Network (EA-Net) is developed in this paper for high-quality video frame interpolation. It follows an end-to-end architecture and can be separated into two stages, \emph{i.e.,} edge-guided flow estimation and edge-protected frame synthesis. In the flow estimation stage, the frame and its extracted edge map are integrated together for flow estimation, so that the flow map are estimated under the guidance of the edge information. Particularly, three edge-aware mechanisms are developed for frame and edge integration, including concatenation, augmentation and two-stream, to analyze their influence to the performance. In the frame synthesis stage, a flow refinement module is designed to refine and interpolate the flow map. Then, an attention module is employed to adaptively attend to the bidirectional flow map when synthesizing the intermediate frames. Besides, the adversarial training strategy is conducted by adopting the frame and edge discriminators to further enhance the reality and preserve the boundary of synthesized frames. The experiments are conducted on three benchmark datasets for both the single and multiple frame interpolation tasks. The results demonstrate that the proposed EA-Net can significantly improve the performance by integrating the edge information explicitly. Moreover, it achieves comparable results with state-of-the-arts in a much compact architecture, and requires no pre-trained models and extra annotated data. 

\subsection{Contributions}

In this paper, the contributions of the proposed EA-Net are summarized into three folds:

1) The edge information is integrated into the video frame interpolation task by adding three simple and effective edge-aware mechanisms, which can reduce the interference caused by large motion and occlusions.

2) The flow attention module is designed to adaptively attend to the bidirectional flow maps, so that more accurate motion information is utilized for frame synthesis.

3) The adversarial training strategy is developed by adopting the frame and edge discriminators, which can enhance the reality of frames and the clarity of object boundaries. 

\subsection{Organization}

The rest of this paper is presented in the following structure. In Section \ref{section2}, the related works of video frame interpolation in the literature are reviewed. In Section \ref{section3}, the detailed architecture of the proposed edge-aware networks is presented, including the edge-guided flow estimation and the edge-protected frame synthesis. In Section \ref{section4}, the experiments are conducted and the results are discussed to analyze the insights of edge information in video frame interpolation. Finally, the conclusions are drawn in Section \ref{section5}.  

\section{Related Works}\label{section2}

Video frame interpolation is a long-standing task in video analysis and an important technique to enhance the video quality. Existing approaches cast this task as synthesizing one or more intermediate frames from temporally neighboring frames \cite{niklaus2020softmax,choi2020channel,lee2020adacof}. They can be roughly classified into three folds, \emph{i.e.}, phase-based approaches, kernel-based approaches and flow-based approaches, which are overviewed in the following subsections.

\subsection{Phase-based Video Frame Interpolation}\label{section2.1}

Motion estimation is the key step in the video frame interpolation task \cite{huang2009correlation}. Phase-based approaches estimate the motion in the videos by computing the phase shift information between frames. This kind of approaches are developed based on the assumption that video motion is captured by the phase shift of pixel color, which have shown promising results in view expansion \cite{zitnick2004high,dai2006accurate} and motion magnification \cite{wadhwa2014riesz,yang2017blind}. However, conventional phased-based approaches can only deal with small range of motion information. To remedy this problem, Meyer \emph{et al.} \cite{meyer2015phase} develop a multi-scale pyramid model and adopt a coarse-to-fine strategy to extend the phase shift limitation, where the phase information is propagated across different levels with a bounded shift correction method. Furthermore, the PhaseNet is proposed to integrate deep learning to phase difference computation \cite{meyer2018phasenet}, which is more robust to handle different scenarios. Besides, Fahim \emph{et al.} \cite{DBLP:conf/iciap/ArifAGR19} employ the edge-preserving guided filtering before computing the phase difference, so that the object boundaries in the frames are protected.
\subsection{Kernel-based Video Frame Interpolation}\label{section2.2}
Kernel-based approaches utilize convolutional neural networks to model the video motion estimation and frame interpolation into an end-to-end architecture \cite{reda2018sdc,zhang2007spatio,DBLP:journals/tcsv/ZhangZJWG09}. Specifically, Niklaus \emph{et al.} \cite{niklaus2017video} develop an adaptive convolution model to estimate the convolution kernels. Naturally, large kernels are used for large motion, vice versa. However, large kernels will increase the computation memory significantly. To reduce this problem, a separable-adaptive convolution model is proposed with the assumption that the 2D kernel can be separated into a pair of 1D kernels \cite{niklaus2017video2}. In this case, the memory consumption is reduced. However, the main drawback of kernel-based approaches is that they cannot deal with the motion larger than the kernel size. Facing this problem, Bao \emph{et al.} \cite{bao2019memc} propose MEMC-Net to integrate motion kernel and optical flow together for video frame interpolation. 

 \begin{figure*}[t]
	\centering
	\includegraphics[width=0.88\textwidth]{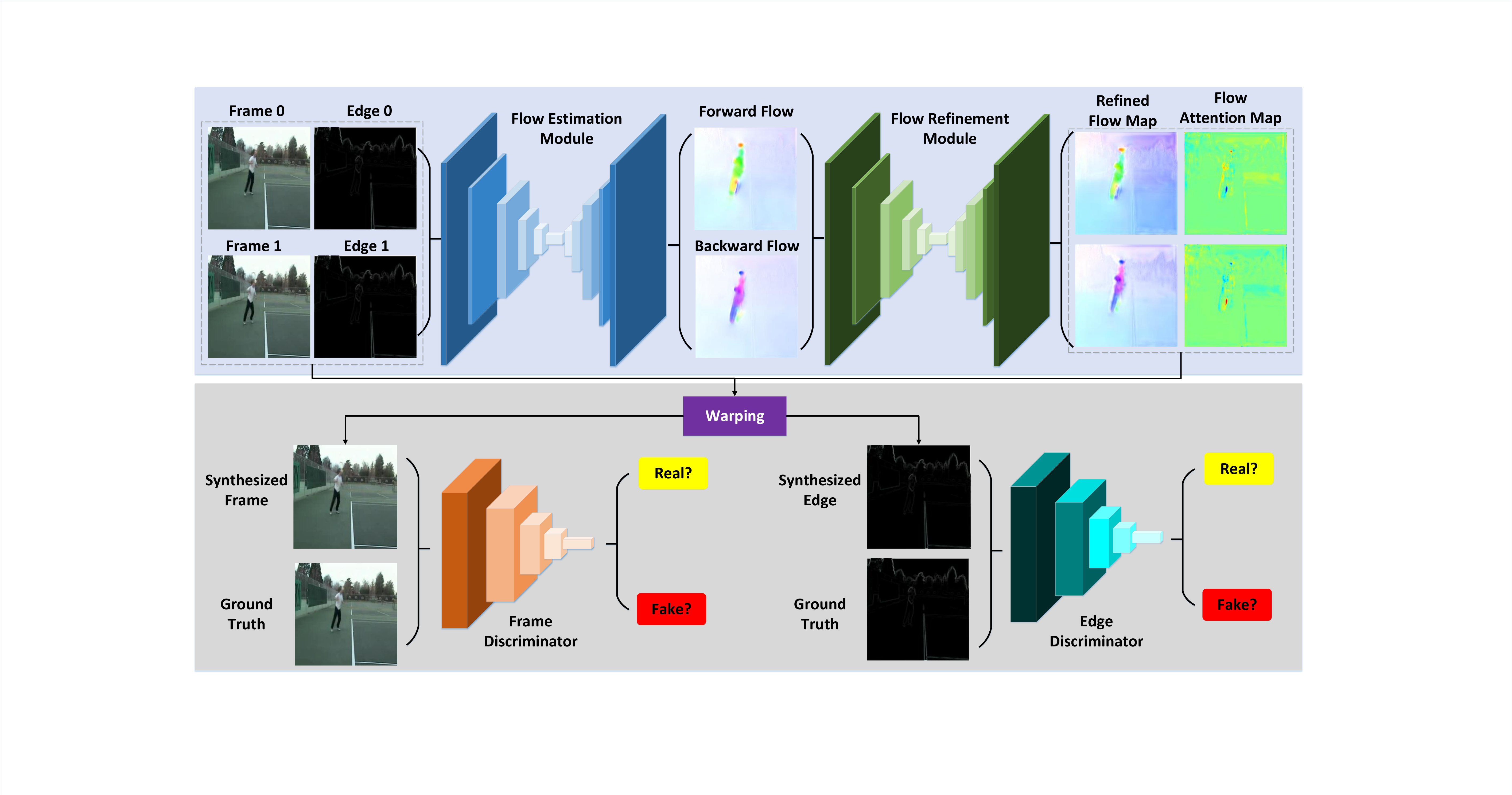}
	\caption{Overview of the proposed EA-Net. It is composed of three parts: 1) The flow estimation module computes the bidirectional flow maps given input frames and their edge maps. 2) The flow refinement module refines the flow maps and generates the bidirectional flow attention maps. 3) The frame and edge discriminators can further enhance the reality and clarity of the synthesized frames.}\label{Fig. 1}
\end{figure*}

\subsection{Flow-based Video Frame Interpolation}\label{section2.3}

Flow-based approaches are most popular in video frame interpolation. This kind of approaches use optical flow to represent the video motion \cite{liu2019deep,li2020video}. Compared with the aforementioned two kinds of approaches, the main advantages of flow-based approaches are that they can generate multiple intermediate frames \cite{jiang2018super,gui2020featureflow}. However, the quality of interpolated frames heavily depends on the accuracy of optical flow. Actually, optical flow computation is a challenging problem in computer vision, since it needs to estimate the pixel-wise correspondence. In this case, conventional optical flow computation methods suffer from serious frame blurriness and artifacts \cite{raket2012motion,werlberger2011optical}. 

To improve the flow accuracy and interpolation quality, Liu \emph{et al.} \cite{liu2017video} propose a deep voxel flow method to learn optical flow with 3D convolutional neural networks, and then generate intermediate frames by trilinear sampling. Jiang \emph{et al.} \cite{jiang2018super} adopt the U-Net architecture for the bidirectional flow computation and multiple frame interpolation. Niklaus \emph{et al.} propose a context-aware video frame interpolation approach, which adopts the PWC-Net \cite{sun2018pwc} as the flow estimator and generates the intermediate frame with a spatial warping layer. Furthermore, Bao \emph{et al.} \cite{bao2019depth} develop a depth-aware approach that integrates the flow, depth, context and kernel together, and interpolates frames with an adaptive warping layer. However, the depth estimation is also a difficult task, and needs pre-training on extra datasets. Lee \emph{et al.} \cite{lee2020adacof} present adaptive collaboration of flows to handle more complex video motion and boost the performance. Although flow-based approaches have made tremendous progress in video frame interpolation, they still suffer from the object boundary blurriness due to the large motion between frames. In this case, we propose a flow-based edge-aware network to preserve the object boundaries in flow estimation and frame interpolation.

\section{Edge-Aware Network for Video Interpolation}\label{section3}
In this paper, we propose an Edge-Aware Network (EA-Net) for the video frame interpolation task. As depicted in Fig. \ref{Fig. 1}, it consists of two stages, \emph{i.e.}, edge-guided flow estimation and edge-protected frame synthesis. Specifically, the edge-guided flow estimation part develops three mechanisms, including edge-augmentation, edge-concatenation and two-stream, to estimate the video motion under the guidance of object boundaries. The edge-protected frame synthesis part adopts a refinement module to refine the flow maps, and an attention module to regulate the weights of forward and backward flow in frame synthesis. Besides, the frame and edge discriminators are adopted to conduct adversarial learning. The detailed architecture of the EA-Net is introduced in the following subsections successively.

\subsection{Edge-guided flow estimation}

To develop an end-to-end architecture, we adopt U-Net \cite{DBLP:conf/miccai/RonnebergerFB15} as the backbone to compute optical flow between two consecutive frames, rather than employ existing flow CNNs, like PWC-Net \cite{sun2018pwc}, FlowNet \cite{DBLP:conf/iccv/DosovitskiyFIHH15}, \emph{etc.}, since they require pre-training on extra datasets. Specifically, U-Net is originally proposed for the medical image segmentation task, which is composed of fully convolutional layers and follows the encoder-decoder architecture. It is quite suitable for the image-to-image task, including image segmentation, image generation as well as flow estimation. In our work, U-Net is employed to compute the video motion from scratch. It should be noted that the flow map is learned in an unsupervised manner and no extra annotated datasets are required in this process. 

Actually, U-Net has been modified as a general network with specific structures for different tasks. In our work, the U-Net adopted for flow estimation is composed of an encoder and a decoder, where the encoder is utilized to extract the high-level frame feature, and the decoder is employed to estimate the motion information given the encoded frame features as input. The encoder contains six down-convolution blocks. Specifically, the filter sizes in the first two layers are set as 7x7 and 5x5, since large receptive fields are helpful to capture the long-range motion, and the rest layers are all with 3x3 filters. Each block consists of two convolutional layers with Leaky ReLU as the activation functions. Except the last block, they are followed by the 2x2 max pooling layer to down-sample the feature map. The decoder consists of 5 up-convolution blocks, which has symmetric structures with the encoder. Specifically, in each block, the only difference is that the pooling layer is replaced by the up-sampling layer, so that the feature maps are up-sampled to the same size with the input frame. Finally, given the consecutive frames as input, the U-Net can generate the bidirectional flow maps correspondingly.

Generally, the accuracy of flow map is essential to the performance of frame interpolation. To better preserve the frame details, we pay special attention to the motion at the object boundaries. In this case, edge-aware mechanisms are developed in the flow estimation process. Firstly, given frame \(I\) as input, the edge map is generated by the canny edge detection algorithm \cite{canny1986computational}, denoted as \(E\). Based on this, the  edge-aware mechanisms are developed as follows.

1) \textbf{Edge-Augmentation.} The object boundaries are emphasized in the original frame by augmenting the pixel values at edges. It is formulated as follows:
\begin{equation}{I^{aug}} = \frac{1}{2}\left( {I + I \odot E} \right),\end{equation}
where $\odot$ stands for the pixel-wise multiplication. The above equation means the non-edge pixels are depressed in the flow estimation process, and the edge pixels are augmented correspondingly.

2) \textbf{Edge-concatenation.} The original frame and the extracted edge map is concatenated as a  six-channel input image. It is formulated as:
\begin{equation}{I^{con}} = \left[ {I ; I \odot E} \right],\end{equation}
where \([ \cdot {\kern 1pt} \,;\, \cdot ]\) denotes the channel-wise concatenation operation. In this case, the number of input channels of U-Net is changed according to \(I^{con}\).

3) \textbf{Two-stream.} A two-stream structure is developed for U-Net, where the flow is estimated by the frames and their edge maps jointly, denoted as:
\begin{equation}F = \frac{1}{2}\left( {{F^I} +  {F^E}} \right),\end{equation}
where \(F^I\) and \(F^E\) stand for the flow map computed by the frame and its edge map, respectively. \(F\) is the finally generated flow map in the two-stream structure.


\subsection{Edge-protected frame synthesis}

After the flow map is computed, the intermediate frames can be simply synthesized. Specifically, given two consecutive frames \(I_0\) and \(I_1\), the intermediate frame \(I_{t}\) can be synthesized by
\begin{equation}{I_t} = warp\left( {{I_0},{F_{t \to 0}}} \right)\; or\; {I_t} = warp\left( {{I_1},{F_{t \to 1}}} \right),\end{equation}
where the bilinear interpolation is employed as the warping function, denoted as \(warp(\cdot)\). \(F_{t \to 0}\) and \(F_{t \to 1}\) are the interpolated flow map. Based on the assumption that the objects are moving uniformly in a small time interval, the intermediate flow maps are computed by 
\begin{equation}{F_{t \to 0}} = t{F_{1 \to 0}} =  - t{F_{0 \to 1}},\end{equation}
\begin{equation}{F_{t \to 1}} =  - \left( {1 - t} \right){F_{1 \to 0}} = \left( {1 - t} \right){F_{0 \to 1}}.\end{equation}

The above describes a simple strategy for intermediate frame synthesis. However, there are remaining two serious problems that can affect the performance: 

1) The flow map is not precise enough to synthesize high-quality intermediate frames, since no supervision is provided in the flow estimator. Additionally, the linear model in Eqn. (5) and (6) works well when the flow changes smoothly, but it is not suitable for other situations, especially at the object boundaries. 


2) There are multiple ways to synthesize the intermediate frames, as depicted in  Eqn. (4), since both the forward and backward flow maps are provided. It is hard to decide which one is better. Practically, the bidirectional flow maps are both important to synthesize the intermediate frame. 


To remedy the above problems, another U-Net network is employed to refine the flow map and synthesize the intermediate frames. It is cascaded after the flow estimator, and they share similar structures. It takes the original frames \(I_0\) and \(I_1\), the estimated flow map \(F_{1 \to 0}\) and \(F_{0 \to 1}\), the interpolated flow map \(F_{t \to 0}\) and \(F_{t \to 1}\), and the warped frames \(warp\left( {{I_0},{F_{t \to 0}}} \right)\) and \(warp\left( {{I_1},{F_{t \to 1}}} \right)\) as inputs. The outputs have two branches, the first is the refined flow map \(F^r_{t \to 0}\) and \(F^r_{t \to 1}\). The second are the attention maps \(A_1\) and \(A_0\), which balance the weights of the forward and backward flow maps in synthesizing the intermediate frame, respectively. Finally, the synthesis of the intermediate frame is formulated as
\begin{equation}{I_t} = {A_0}warp\left( {{I_0},{F^r_{t \to 0}}} \right) + {A_1}warp\left( {{I_1},{F^r_{t \to 1}}} \right),\end{equation}
where the size of \(A_0\) and \(A_1\) is the same with the shape of frames, and 
\begin{equation}{A_0} + {A_1} = \textbf{1}.\end{equation}

After the interpolated frame synthesized, the discriminator is developed to further enhance the reality and clarity of the intermediate frames. The targets of the discriminator are to 1) classify realistic frames from synthesized frames, and 2) classify realistic edge maps from synthesized edge maps. In this case, the discriminator has two branches, \emph{i.e.}, the frame discriminator and the edge discriminator. In this paper, each discriminator is composed of four convolutional layers and one sigmoid layer as the last. In each convolutional layer, the filter size is 4 with stride 2, the initial output channel is 64 and doubles as the layer deepens, and the Leaky ReLU is utilized as the activation function. Besides, the batch normalization layer is stitched after each convolutional layer, which has shown its superiority in reducing the overfitting problem.

\subsection{Optimization}

The final loss function of the proposed EA-Net contains three terms, which are formulated as follows:
\begin{equation}loss = {l^{syn}} + {l^{flow}} + {l^{adv}},\end{equation}
where \(l^{syn}\) is the synthesis loss, \(l^{flow}\) denotes the optical flow loss and \({l^{adv}}\) is the adversarial loss. The details are described as follows.

1) The synthesis loss \(l^{syn}\) measures the similarity of the synthesized frame and the ground truth. In this paper, the similarity is determined by the \(L_1\) norm, which has been proved to be more effective to the interpolation task. It is formulated as:
\begin{equation}{l^{syn}} = {\left\| {{I_t} - I_t^g} \right\|_1},\end{equation}
where \(I_t\) and \(I_t^g\) stand for the synthesized frame and the ground truth, respectively.

2) The flow loss \(l^{flow}\) measures the performance of the estimated flow map. Considering that the annotated flow map is not available, the results of warping frames via the estimated flow map is utilized to evaluate the performance. It is formulated as:
\begin{equation}{l^{flow}} = {\left\| {{I_0} - warp\left( {{I_1},{F_{0 \to 1}}} \right)} \right\|_1} + {\left\| {{I_1} - warp\left( {{I_0},{F_{1 \to 0}}} \right)} \right\|_1}.\end{equation}

3) The adversarial loss \(l^{adv}\) can further constrain the realistic appearance and details of the intermediate frame. It consists of two terms, \emph{i.e.}, the frame loss term and the edge loss term,
\begin{equation}
\begin{array}{l}
l_I^{adv} = {D_I}\left( {I_t^g} \right) - {D_I}\left( {{I_t}} \right)\\
l_E^{adv} = {D_E}\left( {E_t^g} \right) - {D_E}\left( {{E_t}} \right)\\
{l^{adv}} = l_I^{adv} + l_E^{adv}
\end{array}\end{equation}
where \(D_I\) and \(D_E\) represents the frame discriminator and the edge discriminator, respectively.

The proposed EA-Net is conducted on the deep learning platform of PyTorch. In the training procedure, it is optimized with the Adam Optimizer, where the learning rate is initialized as 1e-4, the decay rate is 0.1 with the milestone of 100 epochs. Practically, the convergence can be reached in 500 epochs. 

\section{Experiment}\label{section4}
In the experiment, the proposed EA-Net is evaluated on both the single-frame and multi-frame interpolation tasks, including three benchmarks, \emph{i.e.}, Vimeo90K \cite{xue2019video}, UCF101 \cite{soomro2012ucf101} and Adobe240-fps \cite{su2017deep}. It is compared with several state-of-the-art approaches to verify the effectiveness. Moreover, the ablation studies are conducted to verify the contribution of each part to the frame interpolation performance.

\subsection{Experimental Setup}

\subsubsection{Datasets} 

 \begin{figure}[tp]
	\centering
	\includegraphics[width=0.45\textwidth]{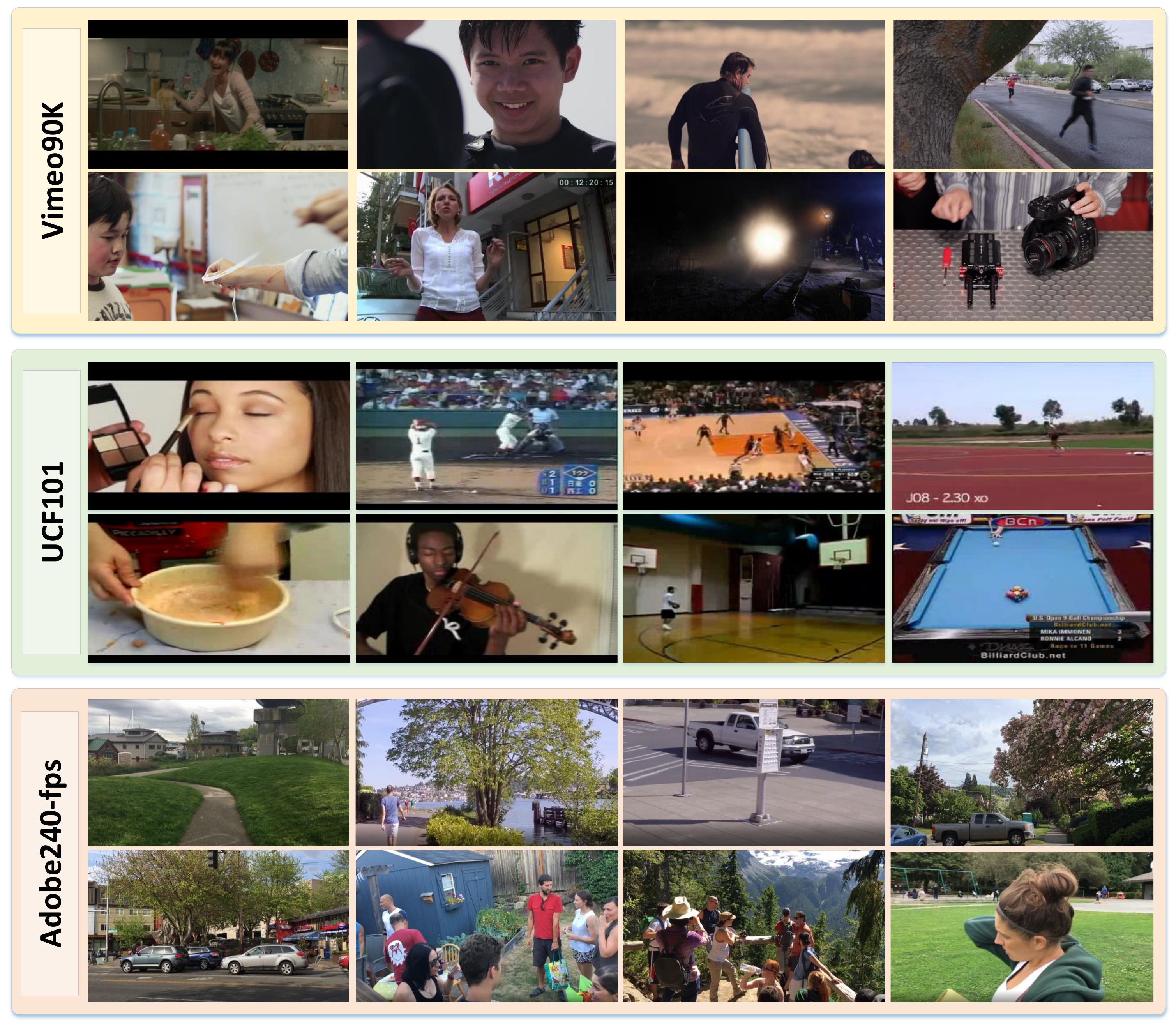}
	\caption{The snapshots of Vimeo90K, UCF101 and Adobe240-fps datasets.}\label{Fig. 2}
\end{figure}

The Vimeo90K dataset \cite{xue2019video} contains 15K video clips downloaded from the vimeo.com\footnote{https://vimeo.com/}.  These video clips are organized as 91701 frame triplets  with a fixed resolution of 448x256. They are employed to perform the single-frame interpolation task, which aims to synthesize the intermediate frame with the left and right frames. The training set is composed of 51313 frame triplets, the test set contains 3782 triplets, and the remaining are used for validation. Besides, in the training process, the data augmentation is conducted, including random frame cropping, frame flipping, and sequence reversing.

The UCF101 dataset \cite{soomro2012ucf101} is a comprehensive video dataset with a variety of human actions. It is conducted for several video based computer vision tasks, including human action recognition, video classification, video prediction, \emph{etc.}.  Following existing protocols, the training process is not conducted on UCF101, and the model trained on Vimeo90K is adopted for evaluation directly. Practically, 379 triplets are utilized for test. They are in the resolution of 256x256.

The Adobe240-fps dataset \cite{su2017deep} is composed of 133 video clips with the frame rate of 240-fps. They are captured by hand-held cameras with the topics in open-domain. These clips contain 79768 frames totally with the resolution of 1280x720. In this paper, the Adobe240-fps dataset is utilized for multi-frame interpolation task. The target is to synthesize the 240-fps video clips from the sampled 30-fps clips. The frames from 112 video clips are used for training, 13 clips for validation, and the remaining 8 clips for test. The data augmentation operation is also conducted for the Adobe240-fps dataset.

\subsubsection{Evaluation metrics} 

Following existing protocols, two metrics are utilized for the evaluation of the video frame interpolation task, including Structural SIMilarity (SSIM) and Peak Signal-to-Noise Ratio (PSNR). They are typical metrics designed for measuring the similarity of two frames. The metrics are positively correlated to the performance. That's to say, higher PSNR and SSIM values mean the synthesized frame and the ground truth are more similar. 

\begin{table*}[ht]
	\centering
	\caption{The results of different approaches on the Vimeo90K and UCF101 datasets.}\label{Table1}
	\renewcommand\arraystretch{1.2}
	
	\begin{tabular}{c|c||p{1.8cm}<{\centering}|p{1.8cm}<{\centering}||p{1.8cm}<{\centering}|p{1.8cm}<{\centering}}
		\hline
		\hline
		&Datasets &\multicolumn{2}{|c||}{Vimeo90K}&\multicolumn{2}{c}{UCF101}\\	
		\hline	
		Approach Type &Approaches &PSNR  &SSIM  &PSNR  &SSIM  \\
		\hline
		Phase-based&Phase-shift \cite{meyer2015phase} &--&-- &30.45 &0.9350  \\\hline
		\multirow {3}*{Kernel-based} 
		&AdaConv \cite{niklaus2017video2} &32.33 &0.9568 &-- &--  \\	
		&SepConv-\(L_f\) \cite{niklaus2017video} &33.45 &0.9674 &34.69 &0.9655  \\
		&SepConv-\(L_1\) \cite{niklaus2017video} &33.79 &0.9702 &34.78 &0.9669\\
		\hline
		\multirow {10}*{Flow-based} 
		&SpyNet \cite{ranjan2017optical} &31.95 &0.9601 &33.67  &0.9633\\
		&Epicflow \cite{ranjan2017optical} &32.02 &0.9622 &33.71&0.9635 \\
		
		&DVF \cite{liu2017video} &31.54 &0.9462 &34.12 &0.9631 \\
		&TOFlow \cite{xue2019video} &33.53 &0.9668 &34.54  & 0.9666 \\
		&Super SloMo \cite{jiang2018super} &32.76 &0.9660 &34.20 &0.9670 \\
		&Super SloMo-meta \cite{choi2020scene} &33.12 &-- &-- &-- \\
		&CyclicGen \cite{liu2019deep} &32.09  &0.9490 &\textbf{35.11}  &\textbf{0.9684}  \\
		&IM-Net \cite{peleg2019net} &33.50  &0.9473 &--  &--  \\
		&MEMC-Net\cite{bao2019memc} &34.29 &0.9739 &34.96 &0.9682\\
		
		&DAIN \cite{bao2019depth} &\textbf{34.71}  &\textbf{0.9756} &\underline{34.99}  &\underline{0.9683}  \\
		\hline
		Proposed&EA-Net &\underline{34.39}	&\underline{0.9753}	&34.97	&0.9675	\\
		\hline
		\hline
	\end{tabular}
	
\end{table*}

 \begin{figure*}[ht]
	\centering
	\includegraphics[width=0.98\textwidth]{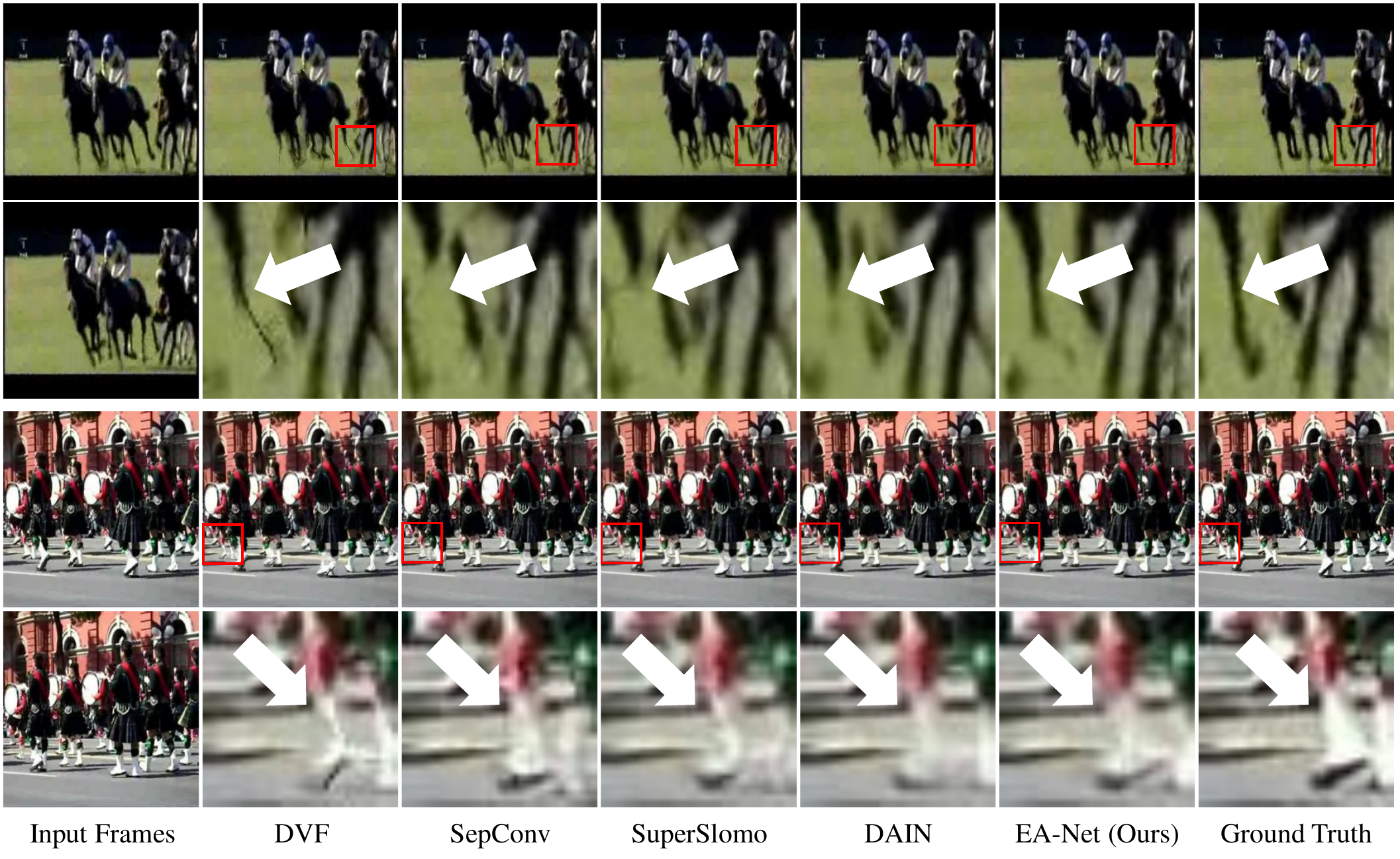}
	\caption{Example results of different approaches on the UCF101 dataset. In each example, the left-most and right-most columns present the input frames and the ground truth of intermediate frames. In the other columns, the first row displays the intermediate frames synthesized by different approaches, and the second row displays the zoom-in details of the region in the red rectangle. }\label{Fig.UCF101_results}
\end{figure*}

\begin{table*}[t]
	\centering
	\caption{Ablation study results on the Vimeo90K and UCF101 datasets.}\label{Table2}
	\renewcommand\arraystretch{1.2}
	
	\begin{tabular}{c|c||p{1.8cm}<{\centering}|p{1.8cm}<{\centering}||p{1.8cm}<{\centering}|p{1.8cm}<{\centering}}
		\hline
		\hline
		&Datasets &\multicolumn{2}{|c||}{Vimeo90K}&\multicolumn{2}{c}{UCF101}\\	
		\hline	
		Module &Modification &PSNR  &SSIM  &PSNR  &SSIM  \\
		\hline
		\multirow {4}*{Edge-aware Mechanism} 
		&Edge-augmentation  &34.39 &0.9753 &34.97 &0.9675  \\	
		&Edge-concatenation &34.34 &0.9751 &34.95 &0.9675  \\
		&Two-stream         &34.35 &0.9746 &34.94 &0.9676\\
		&w/o edge           &33.41 &0.9718 &34.38 &0.9637\\
		\hline
		\multirow {2}*{Attention} 
		&Attention module  &34.39 &0.9753 &34.97 &0.9675  \\	
		&w/o attention module&34.09 &0.9722 &34.84 &0.9640  \\
		\hline		
		\multirow {3}*{Discriminator} 
		&Frame discriminator  &34.18 &0.9714 &34.93 &0.9652 \\
		&Edge discriminator &34.22 &0.9720 &34.71&0.9657 \\
		&w/o discriminator &34.15 &0.9708 &34.61 &0.9651 \\
		
		\hline
		\hline
	\end{tabular}
	
\end{table*}
\subsection{Results on Single-Frame Interpolation}

In this paper, the single-frame interpolation experiment is conducted on the Vimeo90K and UCF101 datasets. In this section, the proposed EA-Net is firstly compared with several state-of-the-arts to show its superiority, and secondly the ablation study is carried out to verify the effectiveness of each component.

\subsubsection{Comparison with state-of-the-arts}

Table \ref{Table1} shows the results of different approaches, including phased-based, kernel-based and flow-based approaches.

Phase-shift is one of the typical phase-based approaches, which proposes a multi-scale pyramid strategy to enhance the capability to deal with the large frame motion. Recently, with the rapid  development of deep learning, traditional phase-based approaches have been surpassed by kernel-based and flow-based approaches. Specifically, AdaConv develops a spatially-adaptive kernel learning method, which can automatically adjust the kernel size according to the estimated local motion. 
SepConv is proposed based on AdaConv by separating 2D kernels into pairs of 1D kernels. In this case, the training difficulty in large motion is reduced, so that SepConv performs better than AdaConv, including both SepConv-\(L_f\) and SepConv-\(L_1\). The proposed EA-Net is developed based on the flow-based architecture. The better performance has demonstrated the superiority of EA-Net in the single-frame interpolation task.

Flow-based approaches are current mainstreams in video frame interpolation. Recently, various of deep convolutional networks are proposed for optical flow estimation. SpyNet proposes a spatial pyramid network to estimate the optical flow. Epicflow has recognized the importance of edge
information for flow estimation, and proposes a sparse-to-dense flow computation scheme, which is more robust to motion boundaries. The better performance of Epicflow than SpyNet indicates the necessity of edge information in video frame interpolation. Our EA-Net outperforms Epicflow significantly. It has verified the superiority of our edge-guided flow estimation and edge-protected frame synthesis. TOFlow is the work constructing the Vimeo90K dataset. It proposes a task-oriented flow computation method, where the optical flow is specially designed for the video interpolation task, so that it gets better performance than previous general approaches, including SpyNet, Epicflow and DVF.

Furthermore, Super SloMo develops a flow interpolation module to synthesize the intermediate frame. The flow refinement module of EA-Net follows similar structures, but the better performance of EA-Net has verified the advantages to consider the edge information in flow computation. CyclicGen introduces the cycle consistency loss and employs the Holistically-nested Edge Detection (HED) algorithm \cite{xie2015holistically} to integrate edge information to synthesize intermediate frames, which requires extra annotations of edge information. However, our EA-Net can perform comparable with CyclicGen on UCF101 and perform significantly better on Vimeo90K, even without extra annotations. To further promote the performance, IM-Net proposes the multi-scale architecture for motion estimation, which can reduce the difficulty in estimating large motion. The edge-guided flow estimation module in EA-Net is an alternative way to address this problem, and the results on the Vimeo90K dataset have verified its superiority.

MEMC-Net and DAIN are two comprehensive approaches. They are both combined with several branches to compute different frame information, including motion, context, kernel, depth and mask. In this case, they achieve the state-of-the-arts on the single-frame interpolation task. However, our EA-Net is an end-to-end approach, which is trained from scratch and doesn't require any pre-trained feature extractors, depth estimators, \emph{etc.}. Although EA-Net is in such a simple architecture, we can see that it performs comparable with MEMC-Net and DAIN. 

To better compare the results, the intermediate frames synthesized by different approaches are displayed in Fig. \ref{Fig.UCF101_results}. Particularly, the results are selected from the frames with large motion, like horse riding and parade. From the zoom-in details of the synthesized frames, it can be clearly observed that EA-Net can better preserve the object boundaries. Overall, the results in Fig. \ref{Fig.UCF101_results} and Table \ref{Table1} can jointly demonstrate the superiority of the proposed EA-Net in the single-frame interpolation task.

\subsubsection{Ablation studies}

 \begin{figure*}[ht]
	\centering
	\includegraphics[width=0.98\textwidth]{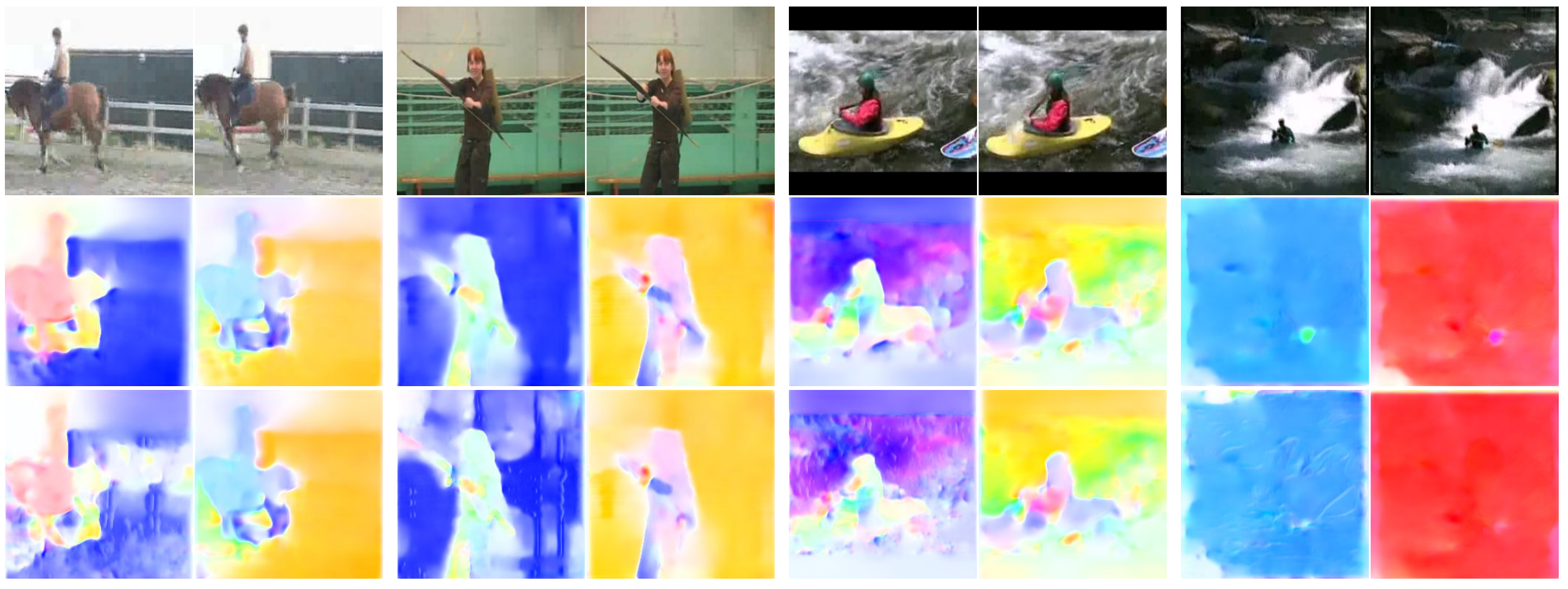}
	\caption{Examples of estimated flow maps. In each two columns, the first row is the input frames. The second row displays the forward and backward flow maps estimated by EA-Net without edge-aware mechanisms. The third row presents the flow maps estimated by EA-Net with the edge-augmentation operation. }\label{Fig.UCF101_flow}
\end{figure*}

Table \ref{Table2} presents the results of ablation studies on the Vimeo90K and UCF101 datasets. The contributions of three components of EA-Net are analyzed, including the edge-aware mechanism, attention module and the discriminator. For the edge-aware mechanism, the results of four baselines are provided. It can be observed that EA-Net without the edge-aware mechanism performs worse than the other three, which meets our motivation and verifies the necessity to integrate the edge information for flow estimation. To better understand the results, the examples of estimated flow maps are displayed in Fig. \ref{Fig.UCF101_flow}. Specifically, in each example, the second row depicts the flow maps estimated by EA-Net without the edge-aware mechanism. The third row is obtained by EA-Net with edge-augmentation operation. It can be seen that the accuracy of the flow maps are promoted by the edge-aware mechanism, especially at the boundaries. That's why EA-Net with edge-aware mechanisms can outperform the one not. Besides, it can be seen from Table \ref{Table2} that EA-Net performs comparably with the three edge-aware mechanisms, which shows its robustness. Closely, we can see that the edge-augment operation performs slightly better than the other two. Thus, it is employed as the default edge-aware mechanism for other comparisons.

 \begin{figure*}[htp]
	\centering
	\includegraphics[width=0.98\textwidth]{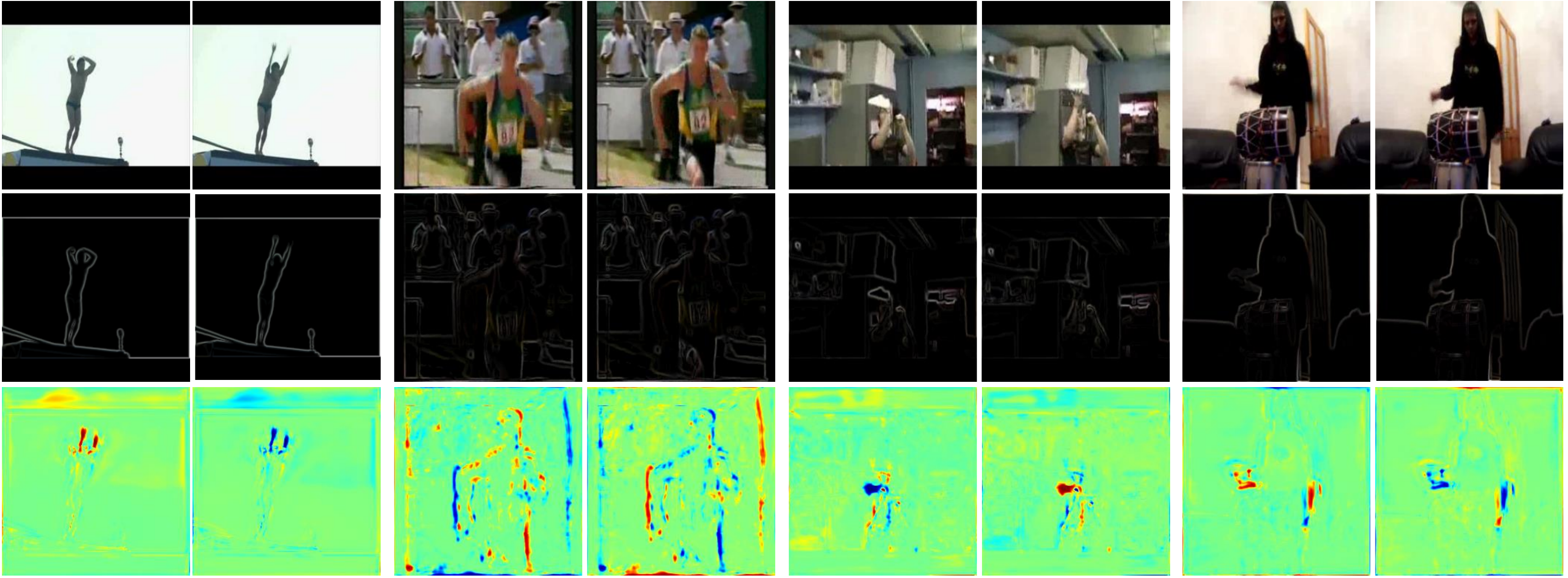}
	\caption{Examples of the edge maps and bidirectional flow attention maps. In each two columns, the first row is the input frames. The second row displays the computed edge maps of corresponding frames. The third row presents the attention maps of the forward and backward flow, respectively. }\label{Fig.UCF101_attention}
\end{figure*}

Table \ref{Table2} presents the comparison of EA-Net with or without the attention module. It should be emphasized that the EA-Net without attention module means the values in \(A_0\) and \(A_1\) are all 0.5. That is to say, the warping results from the forward and backward flow maps are equally contributed to the final interpolated frames. It can be clearly seen that the attention module can significantly boost the performance. To better understand the results, the attention maps are displayed in Fig. \ref{Fig.UCF101_attention}, we can see that the attention weights distribute evenly in the background and are sensitive to the object boundaries obviously, which also illustrates the effectiveness of the proposed edge-aware mechanisms. In this case, EA-Net can better preserve the object boundaries.

Table \ref{Table2} compares the effects of the edge and frame discriminators to the interpolation performance. We can see that the performance gets worse by removing each discriminator. It has quantitatively verified the contributions of the proposed edge and frame discriminators to the frame interpolation performance. Overall, the ablation studies in Table \ref{Table2} have verified the effectiveness of the edge-aware mechanisms in flow estimation, the attention module in frame synthesis, and the discriminators in improving the reality and clarity of synthesized frames.  

\begin{table*}[htp]
	\centering
	\caption{The results of different approaches on the Adobe240-fps dataset.}\label{Table3}
	\renewcommand\arraystretch{1.2}
	
	\begin{tabular}{p{3.8cm}<{\centering}|p{2.8cm}<{\centering}||p{2.8cm}<{\centering}}
		\hline
		\hline
		
		Approaches &PSNR  &SSIM\\
		\hline
		Super SloMo \cite{jiang2018super} &27.52 &0.8593  \\
		Flawless SlowMotion \cite{jin2019learning} &29.24  &0.8754\\
		MEMC-Net \cite{bao2019memc} &30.83 &0.9128 \\
		DAIN \cite{bao2019depth} &\textbf{31.03}  &\underline{0.9172}  \\
		\hline
		EA-Net &\underline{31.02}	&\textbf{0.9205} 	\\
		\hline
		\hline
	\end{tabular}
	
\end{table*}

\begin{table*}[htp]
	\centering
	\caption{Ablation study results on the Adobe240-fps datasets.}\label{Table4}
	\renewcommand\arraystretch{1.2}
	
	\begin{tabular}{c|c||p{1.8cm}<{\centering}|p{1.8cm}<{\centering}||p{1.8cm}<{\centering}|p{1.8cm}<{\centering}}
		\hline
		\hline
		&Refinement Module &\multicolumn{2}{|c||}{with Refinement Module}&\multicolumn{2}{c}{w/o Refinement Module}\\	
		\hline	
		Module &Modification &PSNR  &SSIM  &PSNR  &SSIM  \\
		\hline
		\multirow {4}*{Edge-aware Mechanism} 
		&Edge-augmentation  &31.02 &0.9205 &30.89 &0.9178  \\	
		&Edge-concatenation &30.97 &0.9199 &30.90 &0.9178  \\
		&Two-stream &30.99 &0.9192 &30.88 &0.9167\\
		&w/o edge &30.78 &0.9138 &30.37 &0.9086\\
		\hline
		\multirow {2}*{Attention} 
		&Attention module  &31.02 &0.9205 &30.89 &0.9178 \\	
		&w/o attention module&30.83 &0.9157 &30.68 &0.9122  \\
		\hline		
		\multirow {3}*{Discriminator} 
		&Frame discriminator &30.96 &0.9172 &30.77  &0.9163\\
		&Edge discriminator &30.98 &0.9175 &30.71&0.9166 \\
		&w/o discriminator &30.89 &0.9168 &30.73 &0.9159 \\
		
		\hline
		\hline
	\end{tabular}
	
\end{table*}

\begin{figure*}[t]
	\centering
	\includegraphics[width=0.98\textwidth]{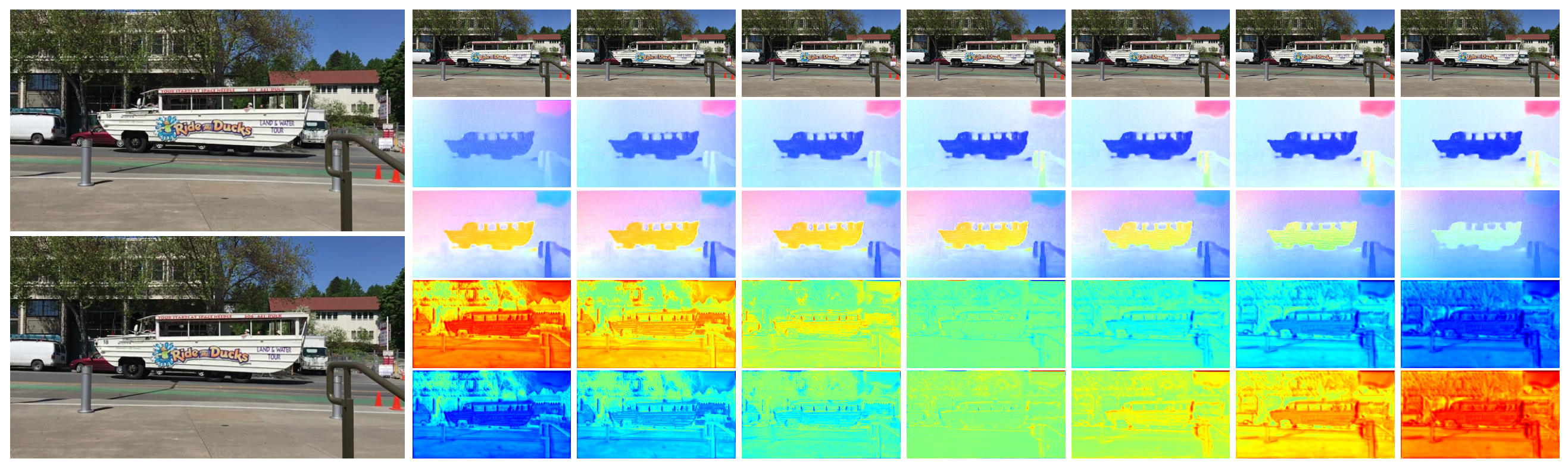}
	\caption{Example results of EA-Net on the Adobe240-fps dataset. The left-most column are the input two frames, \emph{i.e.}, frame 0 and 8. In other columns, the first row displays the intermediate frames synthesized by EA-Net, \emph{i.e.}, frame 1-7. The second and third rows are the interpolated forward and backward flow maps, and the fourth and fifth rows are their corresponding attention maps. }\label{Fig.Adobe240fps_results}
\end{figure*}
\subsection{Results on Multi-Frame Interpolation}

In this paper, the multi-frame interpolation experiment is conducted on the Adobe240-fps dataset. The results of several state-of-the-arts are compared, and the ablation studies are analyzed successively.

\subsubsection{Comparison with state-of-the-arts}

Table \ref{Table3} shows the results of several state-of-the-arts on Adobe240-fps. It should be noted that single-frame interpolation is a special case of the frame interpolation task, where the number of intermediate frames is fixed as one, and the frame rate can only be doubled. As a result, most compared approaches in the single-frame interpolation task can not be generalized to the multi-frame interpolation task directly, where the arbitrary-time intermediate frame synthesis and flow interpolation are required. In this case, only those approaches that are applicable to the multi-frame interpolation task are compared in Table \ref{Table3}.

Four typical multi-frame interpolation approaches are compared in Table \ref{Table3}. Specifically, Super SloMo is a strong baseline in the multi-frame interpolation task. The architecture of the proposed EA-Net is also inspired from it. The main difference of EA-Net and Super SloMo is in the edge-aware processing of frames, including the edge-guided flow estimation and the edge-protected frame synthesis. The significantly better performance of EA-Net than Super SloMo has demonstrated the necessity of taking the edge information into consideration in the multi-frame interpolation task. To better estimate the motion in the video, Flawless SlowMotion separates the interpolation task into two parts, including DeblurNet and InterpNet. They are utilized to deblur the key frames and generate the intermediate frames, respectively. The better performance of Flawless SlowMotion compared with Super SloMo indicates the importance of frame deblurring to the interpolation task. Actually, the proposed EA-Net performs in a similar manner by integrating the edge information into the interpolation task. The better performance of EA-Net has verified the superiority of edge information than frame deblurring.

MEMC-Net and DAIN share similar structures. They combine four encoder-decoder networks to jointly capture different frame information, and another one to synthesize the intermediate frames. Although better results are obtained, they require lots of models pre-trained on other large-scale datasets, which increases the training difficulty and restricts their generality. Fortunately, the proposed EA-Net follows a much simple architecture, where only two encoder-decoder networks are employed, and no extra pre-trained models or annotated data are required. Moreover, the EA-Net surpasses the performance of MEMC-Net and DAIN. The results have shown EA-Net is a simple but effective approach in the multi-frame interpolation task. 

\subsubsection{Ablation studies}
Table \ref{Table4} shows the results of ablation studies on the Adobe240-fps dataset. Similar to the single-frame interpolation task, three components of EA-Net are analyzed, including the edge-aware mechanisms, attention module and discriminators. Furthermore, to verify the effectiveness of the flow refinement module, each baseline is provided two results, \emph{i.e.}, with or without the refinement module. 

From Table \ref{Table4}, it can be observed that the edge-aware mechanisms can significantly promote the performance, and the edge-augmentation operation is slightly better than the other two, which is similar to the results in the single-frame interpolation task. From the displayed flow maps in Fig. \ref{Fig.Adobe240fps_results}, we can see that the object boundaries are very clear in the estimated flow maps. It is benefited from our edge-augmentation operation. Besides, the variance of the bidirectional flow maps according to time are in consistent with the frame motion. It indicates the effectiveness of the flow refinement module, since it can jointly refine and interpolate the flow map.

In Table \ref{Table4}, the attention module can also promote the multi-frame interpolation performance. Actually, it can automatically regulates the weight of the forward and backward flow map in synthesizing the intermediate frames. From the attention maps as depicted in Fig. \ref{Fig.Adobe240fps_results}, we can see that the attention weights vary according to time. On the one hand, the attention weights are negatively correlated to the intervals between the intermediate frame and the input frames. Concisely, when the intermediate frame is near frame 0, the forward flow map is emphasized, and attention weights of the backward flow map are lower, vice versa. It is because the shorter interval between two frames indicates smaller frame motions, and the accuracy of flow maps is more likely to be higher. In this case, the attention module can further reduce the interference caused by the low-quality flow maps. On the other hand, the attention maps are clear at the object boundaries, so that the edges in synthesized frames are better preserved. Besides, in Table \ref{Table4}, the results of EA-Net with both the frame and edge discriminators are better than the other baselines. It is because that the discriminators can promote the reality and clarity of synthesized frames, so that the artifacts and blur are reduced.

Furthermore, in Table \ref{Table4}, the results of EA-Net with and without the refinement module are compared. It can be seen that the refinement module improves the interpolation results significantly, which has verified its necessity. Besides, by comparing the results in Table \ref{Table3} and \ref{Table4}, we can see that EA-Net without the refinement module performs comparably with MEMC-Net and DAIN. Actually, EA-Net without refinement module just contains one encoder-decoder network. It is much more compact than MEMC-Net and DAIN that contain five encoder-decoder networks. The results also demonstrate the superiority of EA-Net in the multi-frame interpolation task.

\section{Conclusions}\label{section5}
In this paper, we propose an Edge-Aware Network (EA-Net) for the video frame interpolation task. It integrates the edge information for flow map estimation and frame synthesis, so that the blur and artifacts of the synthesized intermediate frames are reduced. Specifically, EA-Net is mainly composed of four parts: 1) the edge-aware flow estimation module can compute the flow map between consecutive frames with the help of edge-aware mechanisms. 2) the flow refinement module can further promote the accuracy of the estimated flow map. 3) the bidirectional flow attention module can adaptively regulate the influence of the forward and backward flow map in frame synthesis. 4) the frame and edge discriminators can restrict the synthesized frame to be similar to the real ones, so that the frame quality is improved. Practically, the experimental results on both the single-frame and multi-frame interpolation task have demonstrated the effectiveness of EA-Net and its superiority by integrating the edge information into the interpolation task.

\bibliographystyle{IEEEtran}
\bibliography{IEEEtrans}

\end{document}